\documentclass[runningheads]{llncs}
\usepackage{graphicx}
\begin{document}
\title{A Refined Deep Learning Architecture for Diabetic Foot Ulcers Detection}
%
%
\author{Manu Goyal\inst{1}\orcidID{0000-0002-9201-1385} \and
Saeed Hassanpour\inst{2}\orcidID{0000-0001-9460-6414}}
\authorrunning{Goyal et al.}
%
\institute{Department of Biomedical Data Science, Dartmouth College, Hanover, NH, USA \email{manu.goyal@dartmouth.edu}\\ \and
Departments of Biomedical Data Science, Computer Science, and Epidemiology, Dartmouth College, Hanover, NH, USA
\email{saeed.hassanpour@dartmouth.edu}\\
\url{https://www.hassanpourlab.com}}
\maketitle              
\begin{abstract}
 Diabetic Foot Ulcers (DFU) that affect the lower extremities are a major complication of diabetes. Each year, more than 1 million diabetic patients undergo amputation due to failure to recognize DFU and get the proper treatment from clinicians. There is an urgent  need to use a CAD system for the detection of DFU. In this paper, we propose using deep learning methods (EfficientDet Architectures) for the detection of DFU in the DFUC2020 challenge dataset, which consists of 4,500 DFU images. We further refined the EfficientDet architecture to avoid false negative and false positive predictions. The code for this method is available at https://github.com/Manugoyal12345/Yet-Another-EfficientDet-Pytorch.

\end{abstract}
\section{Introduction}
Diabetes is a serious and chronic metabolic disease characterized by elevated blood glucose. It can further cause major life-threatening complications such as potential blindness; cardiovascular, peripheral vascular and cerebrovascular diseases; kidney failure; and Diabetic Foot Ulcers (DFU), which can lead to lower limb amputation  \cite{wild2004global}. There is a 15-25\% chance that a diabetic patient will eventually develop DFU, and if proper foot care and exams are not taken, lower limb amputation could result  \cite{aguiree2013idf} \cite{armstrong1998validation}. A diabetic patient often needs periodic check-ups with doctors, continuous expensive medication, and hygienic personal care to avoid further adverse health consequences. Hence, caring for a diabetic patient requires a substantial amount of time, and causes a significant financial burden on the patient’s family and health services providers, especially in developing countries where the cost of treating DFU could be equivalent to 5.7 years of annual income \cite{cavanagh2012cost}.

With limited healthcare systems and the proliferation of Information Communication Technology (ICT), intelligent automated telemedicine systems are often cited as one of the most prominent solutions to address  problems associated with DFU assessment. These telemedicine systems can integrate with current healthcare services to provide more cost-effective, efficient, and quality treatment. These systems can also further improve access to patients from rural and remote backgrounds through the use of the Internet and medical imaging technologies  \cite{atzori2010internet}. 

Recently, there has been a growing influence of computer vision and machine learning algorithms in the diagnosis and prognosis of diseases across various modalities of medical imaging, such as MRI, CT, ultrasound, WSI, and dermatology \cite{ahmad2018semantic},  \cite{tomita2018deep}, \cite{yap2018breast}, \cite{wei2019pathologist}, \cite{goyal2017multi}. In the field of DFU assessment, the deep learning methods are utilized by researchers for the recognition and detection of DFU \cite{goyal2017dfunet}, \cite{cruz2020deep},  \cite{goyal2017fully}, \cite{chino2020segmenting}, \cite{goyal2018robust}, \cite{blanco2020superpixel}, \cite{goyal2020recognition}. 

This paper focuses on using automatic DFU detection on the Diabetic Foot Ulcer Challenge (DFUC) dataset, which consists of 4500-DFU images \cite{cassidy2020dfuc2020}. For this challenge, we propose the use of a recent object detection architecture named EfficientDet \cite{tan2020efficientdet}. Further, we refined the predictions of networks on the test set with a score threshold and removed the overlapping bounding boxes.

\begin{figure}[!t]
	\centering
	\begin{tabular}{cc}
		\includegraphics[width=5cm,height=3.75cm]{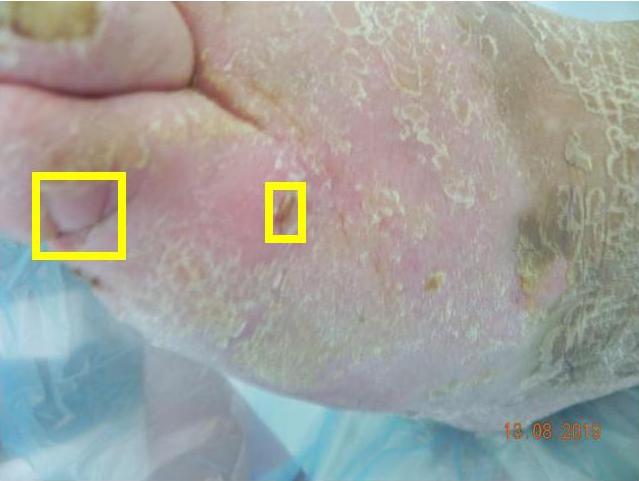} 
		& \includegraphics[width=5cm,height=3.75cm]{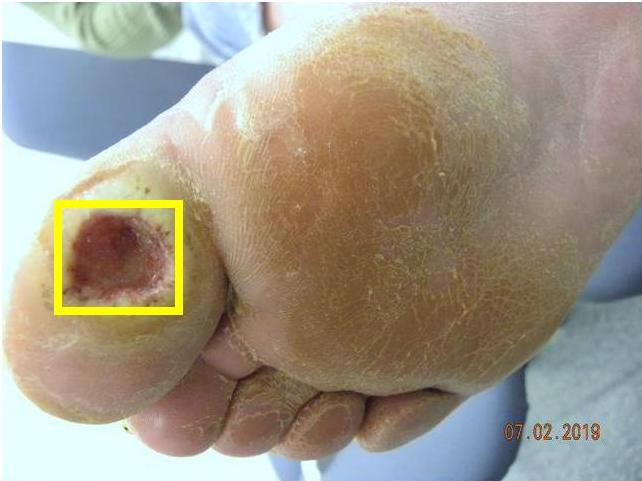} \\
		\includegraphics[width=5cm,height=3.75cm]{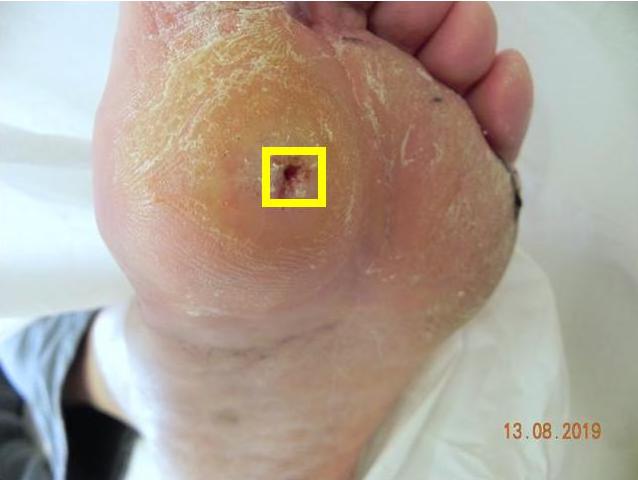} 
		& \includegraphics[width=5cm,height=3.75cm]{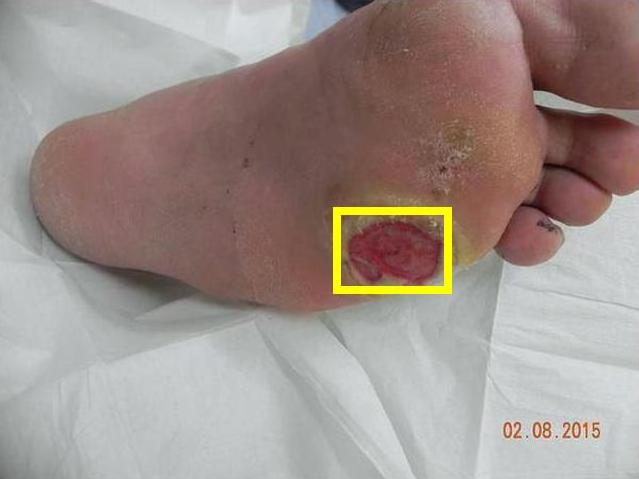} \\
	\end{tabular}     {\tiny }
	
	\caption{Examples of foot images with diabetic foot ulcers with expert annotation}
	\label{fig:exam}
\end{figure}

\section{Methodology}
This section presents a brief description of the DFUC2020 challenge dataset used in this work, pre-processing, data augmentation, and the proposed DFU detection deep learning algorithm.

\subsection{DFUC dataset}
The DFUC2020 dataset consists of 4,500 foot images of DFU collected from the Lancashire Teaching Hospital over the past couple of years.  Three professional cameras (Kodak DX4530, Nikon D3300, and Nikon COOLPIX P100) were used to capture the foot images in this dataset. The foot images with out of focus and blurry artifacts were discarded. The expert annotations were performed by a podiatrist and a consultant physician with specialization in assessing diabetic feet. A few examples of DFU images with annotation are shown in Fig \ref{fig:exam}. This DFUC2020  dataset consists  of  2000 foot images in training, 200  images in validation and 2000 images in the testing set \cite{cassidy2020dfuc2020}. 

\subsection{Pre-Processing}
\begin{figure}[!t]
	\centering
	\begin{tabular}{cc}
		\includegraphics[width=5cm,height=3.75cm]{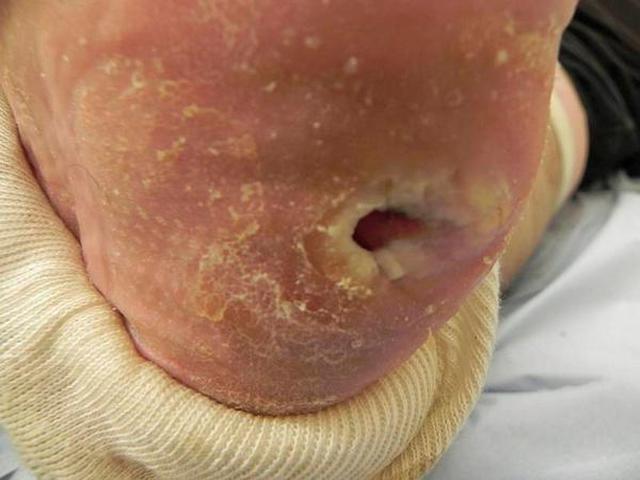} 
		& \includegraphics[width=5cm,height=3.75cm]{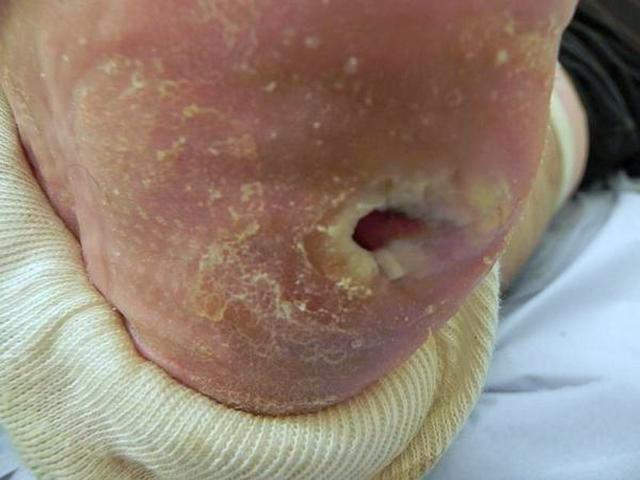} \\
		\includegraphics[width=5cm,height=3.75cm]{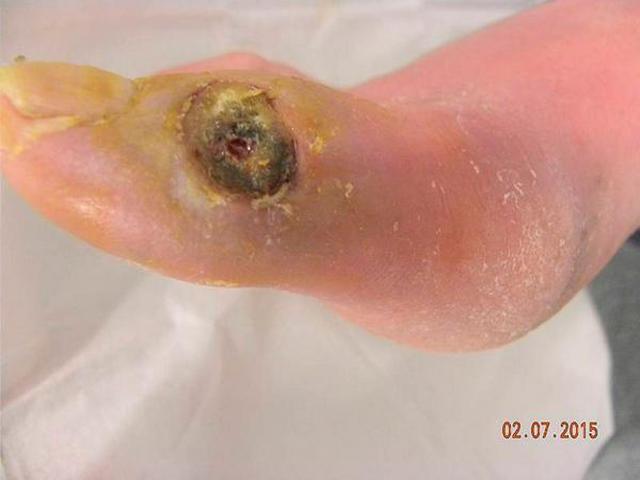} 
		& \includegraphics[width=5cm,height=3.75cm]{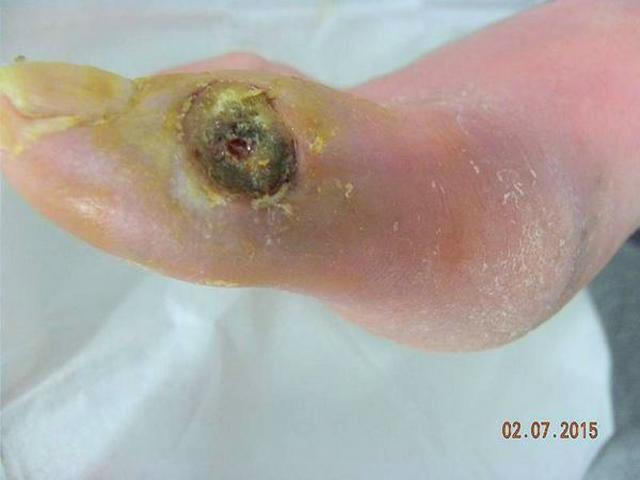} \\
		Original Image
		& After Pre-Processing \\
	\end{tabular}     {\tiny }
	
	\caption{Shades of gray algorithm for pre-processing of DFUC dataset}
	\label{fig:prep}
\end{figure}

Since the dataset was captured with different types of camera devices and lighting conditions, we used the color constancy algorithm named Shades of Gray to cope with noise and lighting effects from different capturing devices \cite{hua2019effect}. The examples of pre-processing of DFU images are shown in Fig. \ref{fig:prep}.

\subsection{Data-augmentation}
\begin{figure}[!t]
	\centering
	\begin{tabular}{ccc}
		\includegraphics[width=4cm,height=3cm]{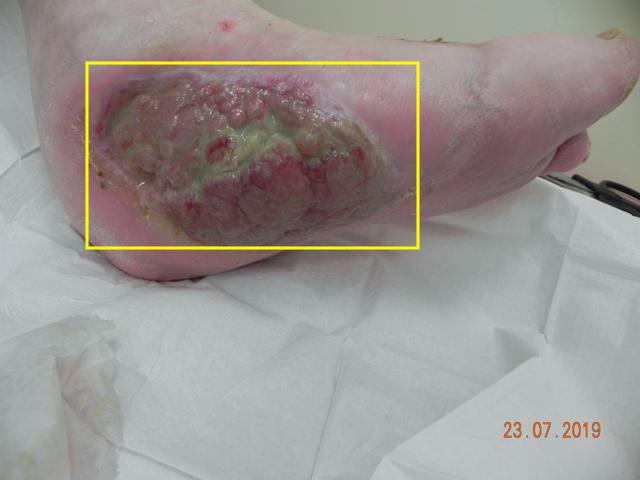} 
		& \includegraphics[width=4cm,height=3cm]{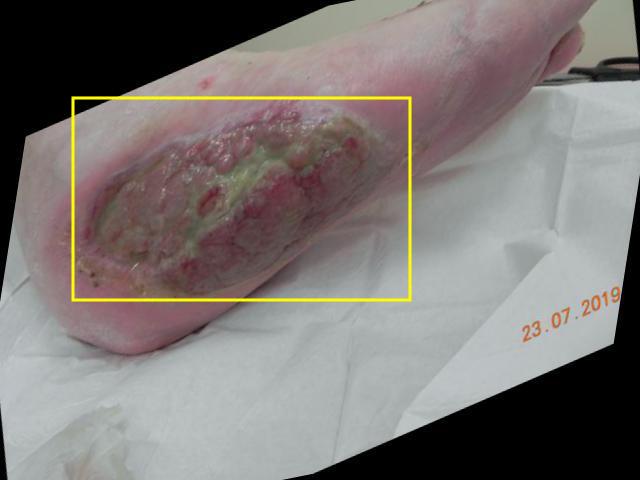} 
		& \includegraphics[width=4cm,height=3cm]{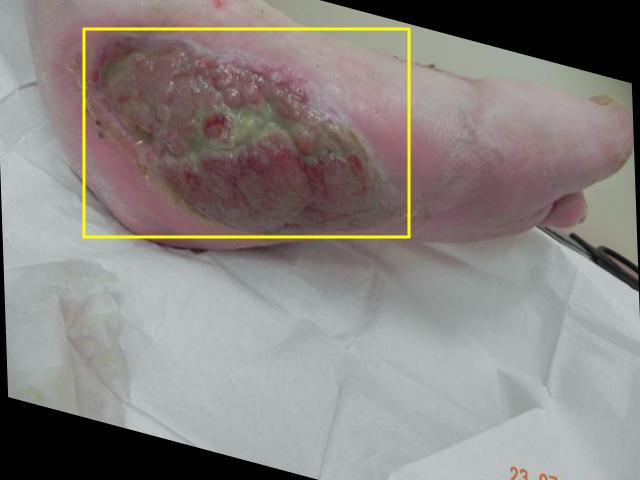} \\
		\includegraphics[width=4cm,height=3cm]{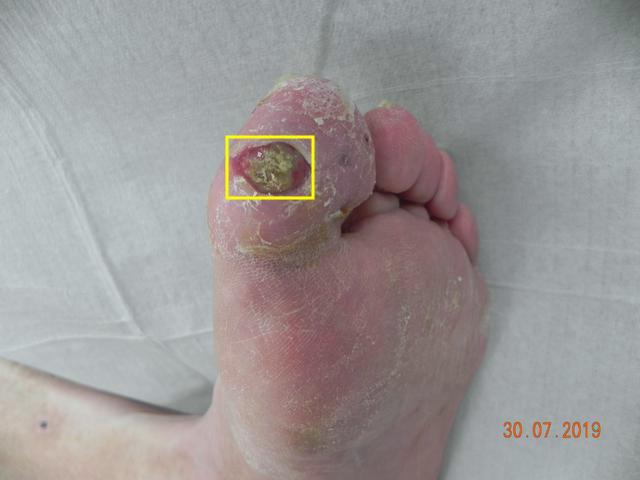} 
		& \includegraphics[width=4cm,height=3cm]{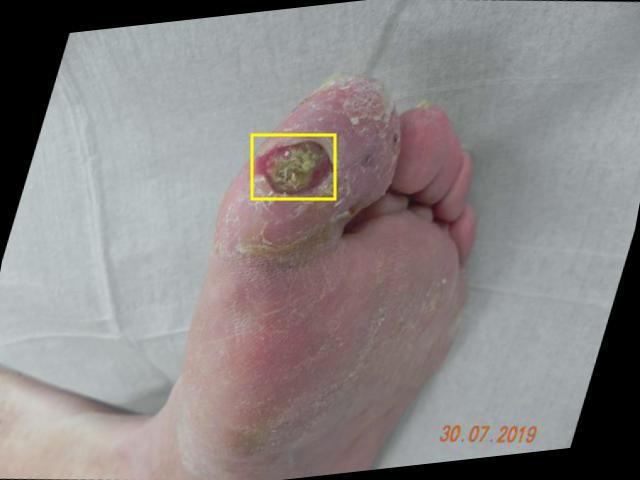}
		& \includegraphics[width=4cm,height=3cm]{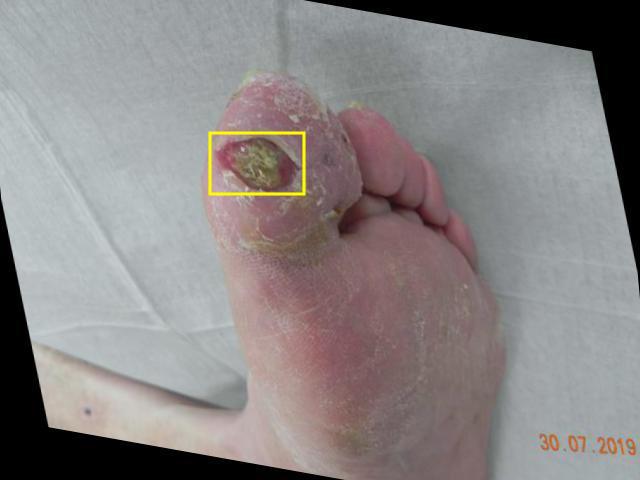}  \\
		Original Image
		& Augmented Image  & Augmented Image \\
	\end{tabular}     {\tiny }
	
	\caption{Shades of gray algorithm for pre-processing of DFUC dataset}
	\label{fig:dataaug}
\end{figure}

Data-augmentation techniques proved to be an important tool in improving the performance of deep learning algorithms for various computer vision tasks \cite{shorten2019survey}, \cite{goyal2018region}, \cite{yap2020breast}. For this task, we augmented the training data by applying identical transformations to the images and associated bounding boxes for DFU detection. We mostly used two common types of transformation, i.e., random rotation and shear, to augment the DFUC dataset. A few examples of data-augmentation for the DFUC dataset are shown in  Fig. \ref{fig:dataaug}.

\begin{figure}[!t]
	\centering
	\begin{tabular}{cc}
		\includegraphics[width=5cm,height=3.75cm]{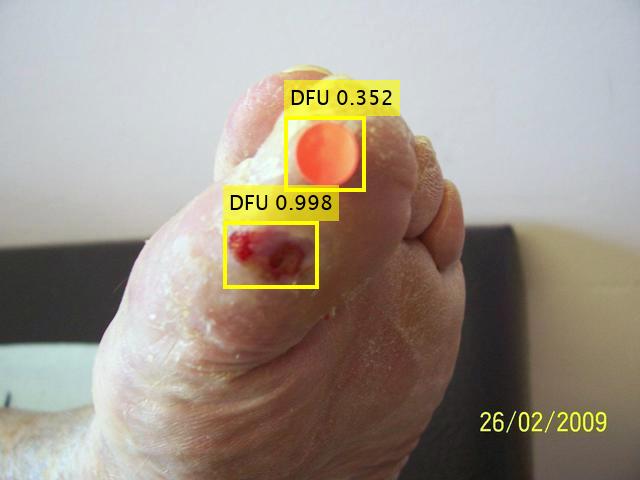} 
		& \includegraphics[width=5cm,height=3.75cm]{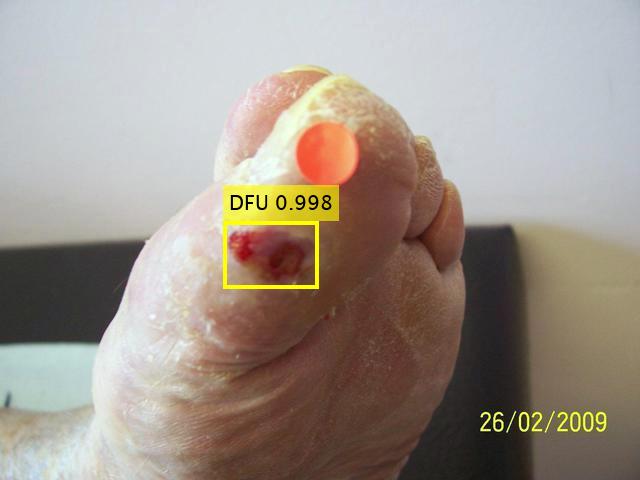} \\
		\includegraphics[width=5cm,height=3.75cm]{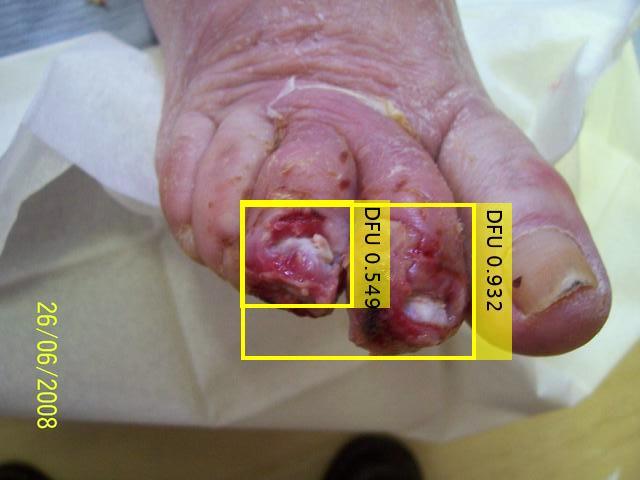} 
		& \includegraphics[width=5cm,height=3.75cm]{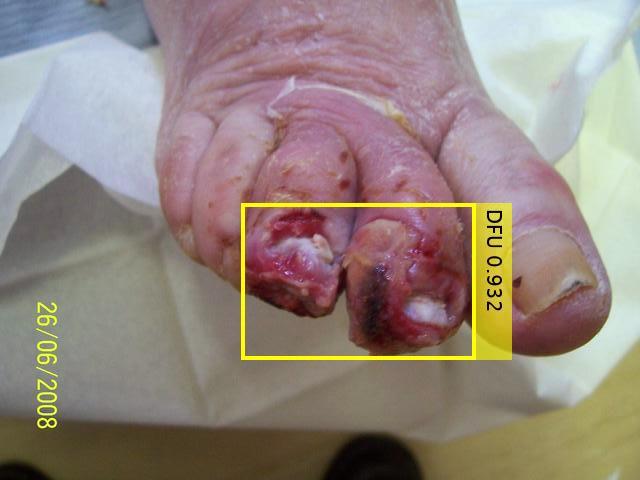} \\
		Inference by original algorithm
		& Final inference by refined algorithm\\
	\end{tabular}     {\tiny }
	
	\caption{Final inference by refined EfficientDet algorithm }
	\label{fig:fp}
\end{figure}

\begin{table*}[]
	\centering
	\addtolength{\tabcolsep}{2pt}
	\renewcommand{\arraystretch}{1.5}
	\caption{Performance of EfficientDet Architectures on MS-COCO dataset}
	\label{Effic}
	\scalebox{0.8}{
		\begin{tabular}{cc} \hline
			Model                                                                                                                                 & Test AP                       \\ \hline \hline
			EfficientDet-DO & 33.8                                          \\ 
			EfficientDet-D1 & 39.6                                          \\ 
			EfficientDet-D2 & 43.0                                          \\ 
			EfficientDet-D3 & 45.8                                          \\ 
			EfficientDet-D4 & 49.4                                          \\ 
			EfficientDet-D5 & 50.7                                          \\ 
			EfficientDet-D6 & 51.7                                          \\ 
			EfficientDet-D7 & 53.7                                          \\ \hline
	\end{tabular}}
\end{table*}

\subsection{EfficientDet Method}

EfficientDet is a new family of object detectors that uses a weighted bi-directional feature pyramid network (BiFPN) and a compound scaling method that uniformly scales the resolution, depth, and width for all backbone, feature network, and bounding box/class prediction networks at the same time \cite{tan2020efficientdet}. EfficientDet achieved state-of-the-art accuracy on a popular object detection dataset called MS-COCO, as shown in Table \ref{Effic}. 

We further refined the EfficientDet architectures with a score threshold of 0.5 and removed overlapping bounding boxes to minimize the number of false positives and false negatives. The scores were compared between the overlapping bounding boxes, and the bounding box with the highest score was used as the final output. The examples of final output by refined EfficientDet architecture is shown in Fig. \ref{fig:fp}.

\section{Experiment and Results} 
We used the EfficientDet codebase (https://github.com/zylo117/Yet-Another-EfficientDet-Pytorch) in PyTorch to design and build an object detection model for DFU. For training, we utilized the pre-trained weights of networks trained on the MS-COCO dataset, which consists of more than 80,000 images of 90 classes, to fine-tune our DFU detection model \cite{huang2016speed} \cite{goyal2017fully}. 

\begin{figure}[!t]
	\centering
	\begin{tabular}{cc}
		\includegraphics[width=5cm,height=3.75cm]{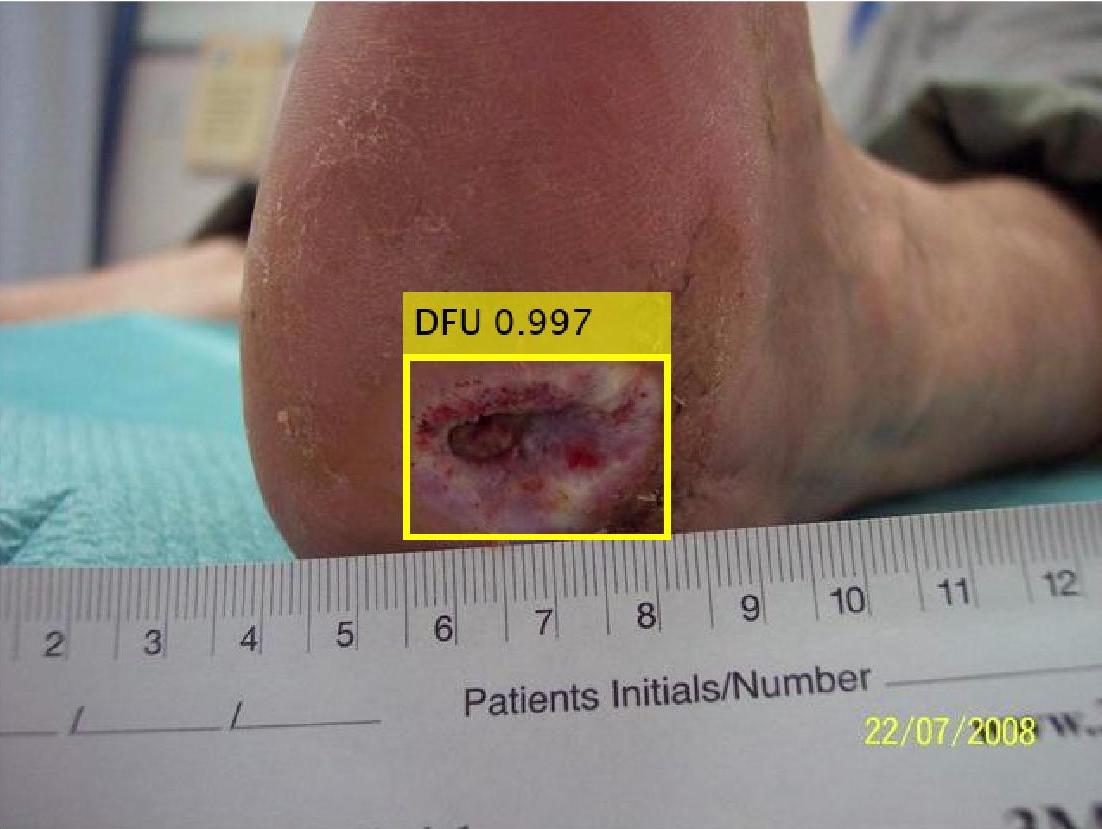} 
		& \includegraphics[width=5cm,height=3.75cm]{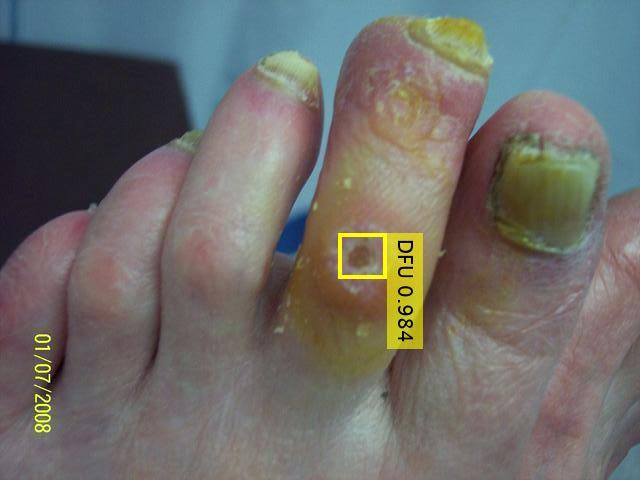} \\
		\includegraphics[width=5cm,height=3.75cm]{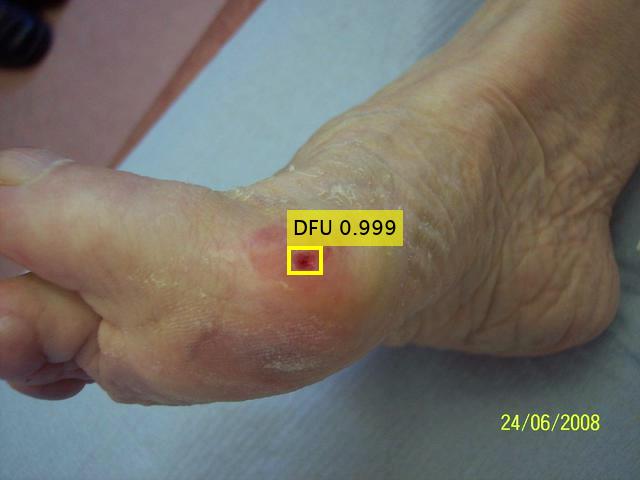} 
		& \includegraphics[width=5cm,height=3.75cm]{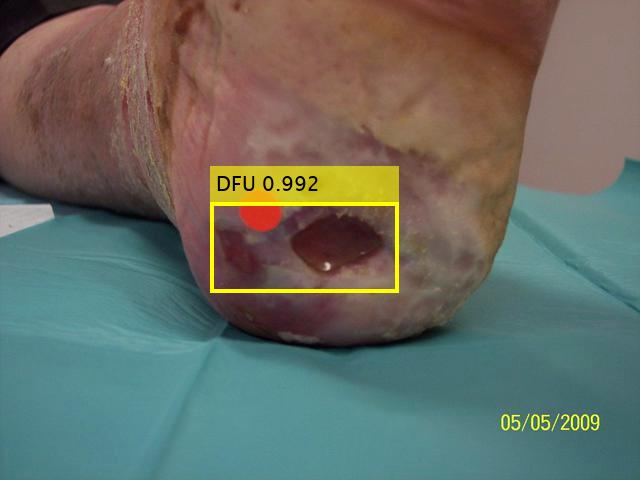} \\
	\end{tabular}     {\tiny }
	
	\caption{Examples of DFU detection by EfficientDet algorithm in testing set}
	\label{fig:inf}
\end{figure}

Since expert annotations were not available for the validation and testing DFUC datasets, we split an original training dataset of the DFUC challenge randomly into 90\% data to train the EfficientDet algorithms, and 10\% data for the validation set. We used the DFUC training set to train the networks on an NVIDIA QUADRO RTX 8000 GPU.  We trained different EfficientDet architectures for 100 epochs, in addition to having an \texttt{Adam\_w} optimizer, with a learning rate of 0.00003, and a decay factor of 0.95. The final model was selected based on minimum validation loss. A few examples of inference on the testing set are shown in Fig. \ref{fig:inf}.

\section{Conclusion}
In this work, we propose the refined EfficientDet algorithms for DFU detection in the DFUC dataset, which consists of 4,500 foot images with ulcers. To improve the performance of the algorithms, we pre-processed the dataset with the Shades of Gray algorithm. We extensively used the data augmentation techniques to learn the subtle features of DFUs of various sizes and grades. We further refined the inference of the original efficientDet method by using a score threshold and removing overlapping bounding boxes. We submitted multiple result files with different models and score thresholds for the DFUC 2020 challenge.
%
%
 \bibliographystyle{splncs04}
 \bibliography{Ref}

\end{document}